\documentclass{article}
\usepackage{spconf,amsmath,epsfig}

\usepackage{booktabs} 
\usepackage{multirow}
\usepackage{caption}
\usepackage{pifont}
\usepackage[table]{xcolor}
\usepackage{subcaption}
\usepackage{amssymb}

\newcommand{\J}{$\mathcal{J}$}
\newcommand{\Jseen}{$\mathcal{J}_\textrm{seen}$}

\newcommand{\Fseen}{$\mathcal{F}_\textrm{seen}$}
\newcommand{\Funseen}{$\mathcal{F}_\textrm{unseen}$}
\newcommand{\F}{$\mathcal{F}$}
\newcommand{\Gsc}{$\mathcal{G}$}
\newcommand{\JandF}{$\mathcal{J} \& \mathcal{F}$}
\newcommand{\yesmark}{\ding{51}}
\newcommand{\nomark}{\ding{55}}
\definecolor{mygray}{gray}{.9}
\definecolor{rouse}{rgb}{0.981,0.961,0.941}
\definecolor{light-yellow}{rgb}{1,1,0.93}
\definecolor{light-green}{rgb}{0.95,1,0.95}
\newcommand{\pub}[1]{\color{gray}{\footnotesize{{#1}}}}

\let\OLDthebibliography\thebibliography
\renewcommand\thebibliography[1]{
  \OLDthebibliography{#1}
  \setlength{\parskip}{0pt}
  \setlength{\itemsep}{0pt plus 0.3ex}
}

\begin{document}\sloppy

\def\x{{\mathbf x}}
\def\L{{\cal L}}

\title{Video Object Segmentation with Dynamic Query Modulation\thanks{$^{*}$ Corresponding author}}

%
\name{Hantao Zhou$^{1}$ $\quad$Runze Hu$^{2*}$$\quad$Xiu Li$^{1*}$}

\address{$^{1}$Tsinghua University$\quad$$^{2}$Beijing Institute of Technology}

\maketitle

\begin{abstract}

Storing intermediate frame segmentations as memory for long-range context modeling, spatial-temporal memory-based methods have recently showcased impressive results in semi-supervised video object segmentation (SVOS). However, these methods face two key limitations: 1)  relying on non-local pixel-level matching to read memory, resulting in noisy retrieved features for segmentation; 2) segmenting each object independently without interaction.  These shortcomings make the memory-based methods struggle in similar object and multi-object segmentation. 
To address these issues, we propose a query modulation method,  termed QMVOS. This method summarizes object features into dynamic queries and then treats them as dynamic filters for mask prediction, thereby providing high-level descriptions and object-level perception for the model.
Efficient and effective multi-object interactions are realized through inter-query attention.
Extensive experiments demonstrate that our method can bring significant improvements to the memory-based SVOS method and achieve competitive performance on standard SVOS benchmarks.
The code is available at https://github.com/zht8506/QMVOS.

\end{abstract}
\begin{keywords}
SVOS, Memory bank, Object query
\end{keywords}
\section{Introduction}
\label{sec:intro}

Video Object Segmentation (VOS) is a foundational yet challenging video understanding task, requiring precise segmentation of specific objects in a video. 
This paper focuses on the semi-supervised VOS, abbreviated SVOS, which aims to accurately and promptly segment the objects throughout the video given only an annotated object mask on the first frame. 
SVOS methods find broad applications in tasks like video editing~\cite{cheng2021modular}, automatic labeling~\cite{bekuzarov2023xmem++}, and video tracking~\cite{cheng2023tracking}.

\begin{figure}[!t]
\centering
\includegraphics[scale=0.45]{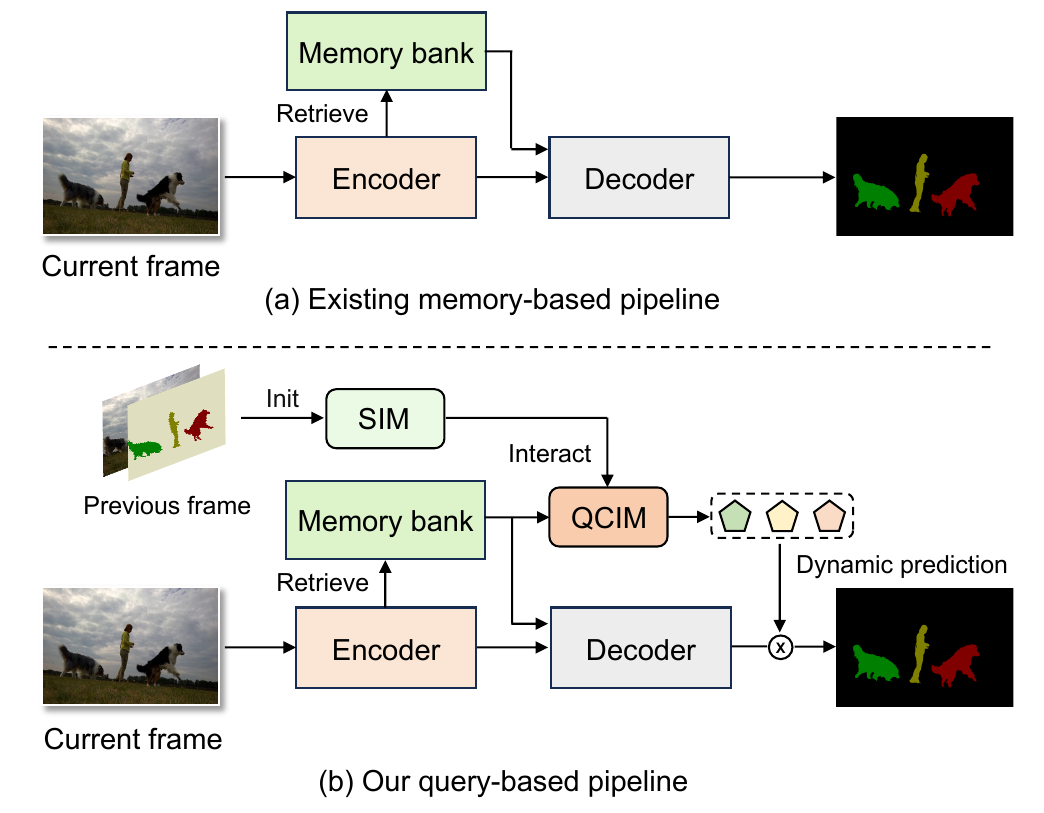}
\caption{The pipeline of existing memory-based SVOS works (a) and our QMVOS (b). Our method innovatively introduces object queries to VOS, enabling effective object-level perception, multi-object interaction and dynamic prediction.}
\label{fig:intro}
\end{figure}

Recently, memory-based paradigms have dominated the SVOS area with leading performance and impressive operating speeds.
A key idea of the memory-based approach is to store segmentation information into memory every few frames and retrieve the memory when a new frame requires segmentation, as depicted in Fig.~\ref{fig:intro}(a). 
This strategy can propagate the predicted frame information, enabling the model to leverage the spatio-temporal information of historical frames.

However, there are two limitations of the existing memory-based methods: \textit{modeling dynamic object-level perception} and \textit{performing multi-object interaction}. 
In particular, memory-based methods employ non-local dense matching to adaptively read valuable segmentation information from the memory bank, based on the similarity between the features of the current frame and those from historical frames.
However, the pixel-level matching is object-agnostic and lacks global perception of the object. The retrieved features may be noisy when the matching process is distracted by the background, leading to inaccurate segmentation~\cite{cho2022tackling}. Another weakness is that existing memory-based methods cannot perform multi-object interaction efficiently. These methods typically segment each object independently and then ensemble all single-object predictions into a multi-object segmentation. However, this approach is ineffective in exploring multi-object contextual information and therefore falls short in segmenting similar objects. While a simple solution involves performing self-attention between the features of each object, it is computationally inefficient and negatively impacts the real-time performance of the model.

In this paper, we present a query-based modulation method, QMVOS, which introduces object-level descriptions to complement the pixel-level memory-based paradigm. 
We represent the objects as compact ``object queries'', a concept initially proposed by DETR~\cite{carion2020end} in object detection and later extended to many other fields~\cite{cheng2022masked,li2023mask,yang2023boxsnake}.
The pipeline of our work is illustrated in Fig.~\ref{fig:intro}(b). We use the predicted mask and features of the previous frame to summarize object queries. These queries will interact with the contents of the current frame. Then, we treat them as dynamic filters to perform dot-products with the features output from the decoder to obtain mask predictions. This paradigm exhibits three advantages: 1) it can provide object-level perception for the object-agnostic memory-based method; 2) it converts the prediction of parameter-fixed SVOS model into a dynamic manner; 3) it enables easy and efficient multi-object interaction.

More specifically, we first initialize the object queries by proposing a novel Scale-aware Interaction Module (SIM). This module utilizes multi-scale features and given or predicted mask of previous frame to initialize the object queries, resulting in robust and scale-aware object representation.  
Then, multi-object interaction is conducted by performing self-attention among these queries in SIM. These object queries will be propagated to current frame for prediction. Due to the temporal continuity, the object queries have a high similarity with the appearance of the objects in the current frame, thus providing effective object prior for the prediction of current frame. We further design a Query-Content Interaction Module (QCIM) to  dynamically update queries based on the content of current frame, thus achieving a more accurate content understanding.
Finally, we leverage these queries to filter background information and generate object masks.
This delivers dynamic object-level predictions for the memory-based SVOS approachs.

Our contributions can be summarized as follows:
\begin{itemize}
\setlength{\itemsep}{0pt}
\setlength{\parsep}{0pt}
\setlength{\parskip}{0pt}
\item {} 
We introduce object-level perception and dynamic prediction into current memory-based SVOS methods.
\item {} We devise a Scale-aware Interaction Module for multi-scale initialization and multi-object interaction, and a Query-Content Interaction Module for adaptive query updating and effective content understanding.
\item {} The proposed approach can significantly improve the
performance of the state-of-the-art (SOTA) memory-based method XMem~\cite{cheng2022xmem} in the standard SVOS dataset, without compromising on inference speed.
\end{itemize}

\section{Related work}
\label{sec:rlt_work}

\subsection{Semi-supervised Video Object Segmentation}

Early SVOS approaches~\cite{caelles2017one,bao2018cnn,luiten2018premvos} focus on fine-tuning pre-trained segmentation networks at test time, enabling the model to ``memorize'' annotated objects for segmentation in subsequent frames. 
Despite promising results, these methods often compromise inference speed, as they necessitate fine-tuning for each video.
Different from fine-tuning-based methods, STM~\cite{oh2019video} stores intermediate frames in a memory bank, which encompasses both spatial and temporal information. The model can then reference these stored frames when required to segment a new frame. As of now, memory-based methods~\cite{cheng2022xmem,cheng2021rethinking,mao2021joint,yang2021associating} dominate the SVOS area due to their superior performance and efficiency.

\subsection{Query-based Method}

Since the emergence of DETR, query-based detectors have achieved remarkable success in object detection. The DETR series~\cite{carion2020end,yao2023dynamicbev} regards object detection as a set prediction problem and can be trained end-to-end. This groundbreaking paradigm has found applications in various domains, such as semantic segmentation~\cite{cheng2022masked,li2023mask} and instance segmentation~\cite{yang2023boxsnake}. 
Despite these impressive breakthroughs, their use in SVOS remains underexplored. 
Utilizing the attention mechanism, successfully applied across various domains~\cite{vaswani2017attention,zhou2023unihead,zhou2023etdnet,yao2023ndc,xiao2023semanticac,xiao2023bridging}, 
we effectively introduce query to model object-level perception and enhance multi-object interactions.

\section{method}
\label{sec:method}

\begin{figure*}[!t]
\centering
\includegraphics[scale=0.5]{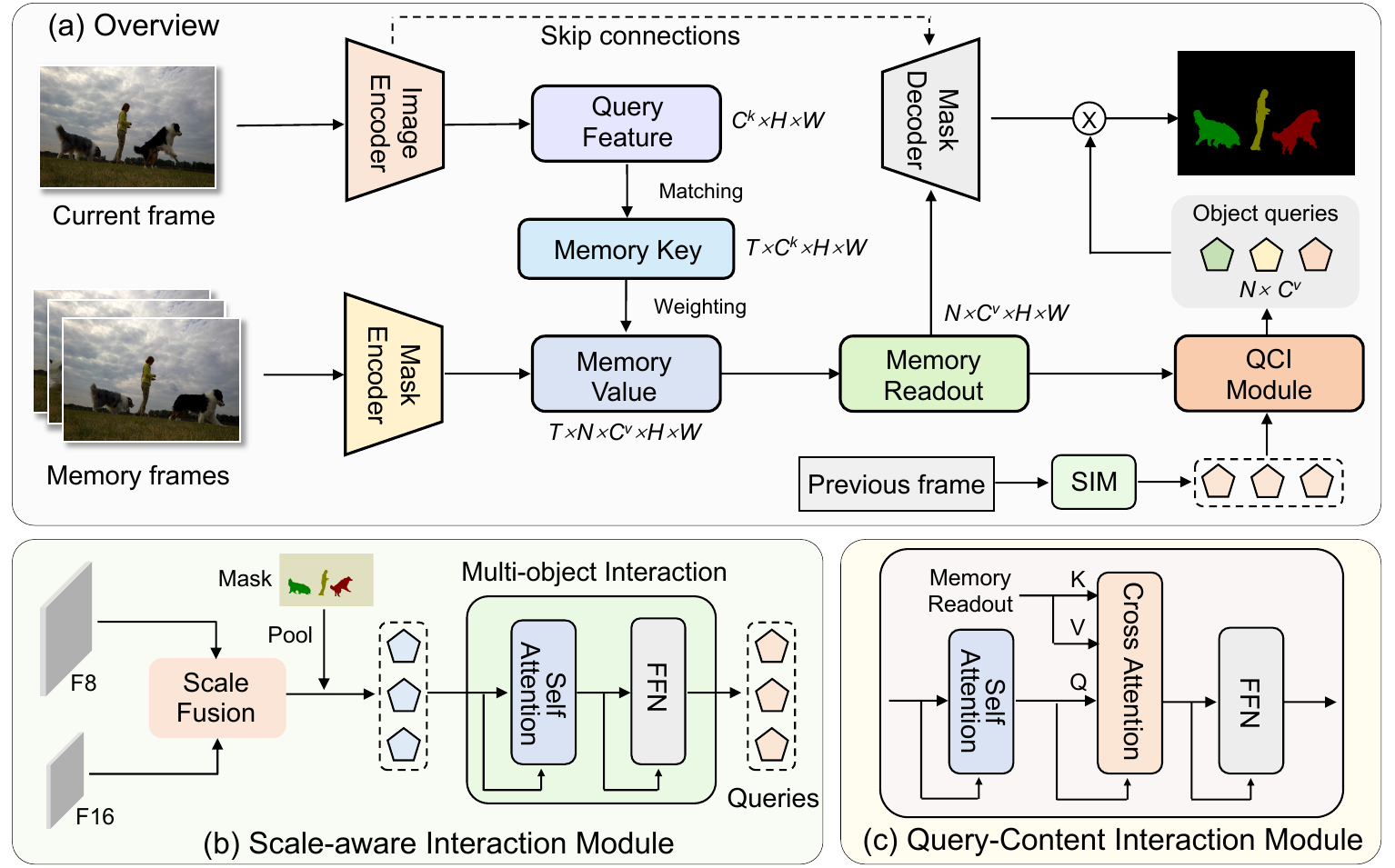}
\caption{(a) The main pipeline of our framework, which follows typical memory-based methods and introduces queries to achieve object-level perception. (b) The structure of SIM. It consists of multi-scale fusion and multi-object interaction processes to initialize queries. (c) The structure of QCIM. It is utilized to perform query-content interaction.}
\label{fig:overview}
\end{figure*}

\subsection{Overview}

The overview of our QMVOS is presented in Fig.~\ref{fig:overview}. 
Following most memory-based methods~\cite{cheng2022xmem,cheng2021rethinking,mao2021joint}, our approach processes frames sequentially and stores the past frames and their predictions into memory. For the current frame that needs to be segmented, the model first extracts hierarchical features through an image encoder. Then, the feature from the  fourth stage ${F_{16} \in \mathbb{R}^ {C^k \times {H} \times {W}}}$ will be used as query feature for memory reading. 
In the memory reading process, the query feature $F_{16}$ is matched with the memory key $M_{K} \in \mathbb{R}^ {T \times C^k \times {H} \times {W}} $ to generate a similarity matrix, which will be used to weight the memory value $M_{V} \in \mathbb{R}^ {T \times N \times C^v \times {H} \times {W}} $. 
Here, $T$ and $N$ denote the number of memory frames and the quantity of objects within the image, respectively. Finally, the memory readout and multi-scale features from the backbone are fed into the mask decoder for mask prediction. 

Different from previous memory-based approaches, our method innovatively employs object queries for mask prediction. These object queries are initialized by the previous frame via the proposed Scale-aware Interaction Module (SIM). Then, we introduce a Query-Content Interaction Module (QCIM) to execute interaction between object queries and readout features, enabling queries to  perceive the content of  the current frame. At last, these queries will perform a dot product operation with the output of the mask decoder to yield the final segmentation mask.

\subsection{Scale-aware Interaction Module}

We propose a Scale-aware Interaction Module (SIM) to initialize the object queries, as depicted in the bottom of Fig.~\ref{fig:overview}(b). SIM leverages multi-scale features and masks of the previous frame to construct high-level object descriptors, called object queries. SIM then employs self-attention to facilitate internal interactions between these queries. In general, SIM can be divided into two steps: scale-aware initialization and multi-object interaction.

\textbf{Scale-aware Initialization.} In this process, we first fuse features of two scales, \textit{i.e.}, features from \mbox{stage-3} $F_{8}$ and \mbox{stage-4} $F_{16}$, to form a feature map containing both fine-grained details and high-level semantic information. Mathematically, the scale fusion can be described as:
\begin{equation}
\begin{aligned}
\hat{F}_{16}&=\mathrm{Conv_1}(\mathrm{Upsample}(F_{16})), \\
F_{fuse}&=\mathrm{Conv_2}(\mathrm{Concat}(\hat{F}_{16},F_{8})), 
\label{eq:scale-fusion}
\end{aligned}
\end{equation}
where $\mathrm{Upsample}$ and $\mathrm{Concat}$ refer to upsampling operation via linear interpolation and channel-wise concatenation, respectively. $\mathrm{Conv_1}$ and $\mathrm{Conv_2}$ represent two $1 \times 1$ convolution layers with corresponding channel number. Then, we utilize multi-scale feature $F_{fuse}$ and mask to generate object queries. Specifically, the mask will be pooled to the same resolution as $F_{fuse}$. Each mask with a specific object is multiplied with $F_{fuse}$. We then employ global average pooling on these masked features to construct object queries.

\textbf{Multi-object Interaction.} Multi-object segmentation in SVOS is very challenging, due to occlusion phenomena and the similarity of multiple objects. A naive method to alleviate this problem is to perform self-attention among the feature of each object. However, this dense and pixel-level  attention approach is computationally inefficient and unfriendly for real-time segmentation applications. 
In contrast, we perform self-attention on these sparse object queries to achieve multi-object interaction in a more efficient manner. 

Specifically, we utilize standard Transformer block~\cite{vaswani2017attention}, which mainly consists of self-attention (SA) and feed forward network (FFN). Let ${X \in \mathbb{R}^ {N \times C^v } }$ denotes the object queries, the multi-object interaction process can be expressed as:
\begin{equation}
\begin{aligned}
\mathrm{SA}(X)&=\mathrm{Softmax}(QK^\top /\sqrt{d})V, \\
\hat{X}&=\mathrm{LN}(\mathrm{SA}(X)+X), \\
X_{sim}&=\mathrm{LN}(\mathrm{FFN}(\hat{X})+\hat{X}), 
\label{eq:self-att}
\end{aligned}
\end{equation}
where $Q$, $K$, $V$ are the learned linear transformation of $X$, representing query, key and value matrices respectively; $d$ is the dimension of $K$; $\mathrm{LN}$ denotes the layer normalization; $\mathrm{FFN}$ consists of two fully connected (FC) layers.

\begin{table}[!t]
\setlength{\abovecaptionskip}{0.1cm} 
\centering
\caption{Quantitative comparisons on the DAVIS~2016~\cite{perazzi2016benchmark} and DAVIS~2017~\cite{pont20172017} val split. OF and M denotes online finetuning and memory-based methods, respectively. $^{*}$ represents the baseline method used in this paper.}
\small
{
\begin{tabular}{lcccccc}
\toprule
\rowcolor{mygray}
Method & Reference  & OF  & M       & \JandF{}  &  \J{} & \F{} \\
\midrule
\multicolumn{7}{c}{\emph{DAVIS 2016 Validation Split}} \\
\midrule
PReMVOS~\cite{luiten2018premvos} & \pub{ACCV18} & \yesmark & \nomark &  86.8 & 84.9 & 88.6 \\
CFBI~\cite{yang2020collaborative} & \pub{ECCV20} & \nomark & \nomark &  89.4 & 88.3 & 90.5 \\
AOT-L~\cite{yang2021associating} & \pub{NeurIPS21} & \nomark & \yesmark &  90.0 & 88.7 & 91.4 \\
TBD~\cite{cho2022tackling} & \pub{ECCV22} & \nomark & \nomark & 84.3 & 85.2 & 83.4 \\
XMem$^{*}$~\cite{cheng2022xmem} & \pub{ECCV22} & \nomark & \yesmark & 90.2 & 89.0 & 91.4 \\
\rowcolor{rouse}
\textbf{QMVOS} & - & \yesmark  &  \yesmark & \textbf{90.5} & \textbf{89.2} & \textbf{91.8} \\
\midrule
\midrule
\multicolumn{7}{c}{\emph{DAVIS 2017 Validation Split}} \\
\midrule
OSVOS~\cite{caelles2017one} & \pub{CVPR17} & \yesmark & \nomark &  60.3 & 56.6 & 63.9 \\
CINM~\cite{bao2018cnn} & \pub{CVPR18} & \yesmark & \nomark &  70.6 & 67.2 & 74.0 \\
PReMVOS~\cite{luiten2018premvos} & \pub{ACCV18} & \yesmark & \nomark &  77.8 & 73.9 & 81.7 \\
LWL~\cite{bhat2020learning} & \pub{ECCV20} & \nomark  & \nomark &  81.6 & 79.1 & 84.1 \\
CFBI~\cite{yang2020collaborative} & \pub{ECCV20} & \nomark & \nomark &  81.9 & 79.1 & 84.6 \\
SST~\cite{duke2021sstvos} & \pub{CVPR21} & \nomark & \nomark & 82.5 & 79.9 & 85.1 \\
STCN~\cite{cheng2021rethinking} & \pub{NeurIPS21} &  \nomark & \yesmark &  82.5 & 79.3 & 85.7 \\
CFBI+~\cite{yang2021collaborative} & \pub{TPAMI21} & \nomark & \nomark &  82.9 & 80.1 & 85.7 \\
JOINT~\cite{mao2021joint} & \pub{ICCV21} & \nomark  & \yesmark &  83.5 & 80.8 & 86.2 \\
AOT-L~\cite{yang2021associating} & \pub{NeurIPS21} & \nomark & \yesmark &  83.6 & 80.5 & 86.6 \\
TBD~\cite{cho2022tackling} & \pub{ECCV22} & \nomark & \nomark & 75.2 & 73.2 & 77.2 \\
XMem$^{*}$~\cite{cheng2022xmem} & \pub{ECCV22} & \nomark & \yesmark & 83.4 & 80.5 & 86.3 \\
\rowcolor{rouse}
\textbf{QMVOS} & - & \yesmark  &  \yesmark   & \textbf{84.9} & \textbf{81.4} & \textbf{88.3} \\
\bottomrule
\end{tabular}
\label{tab:sota-davis}
}
\end{table}

\subsection{Query-Content Interaction Module}
\label{sec:QIM}

The queries initialized from the previous frame contain similar object information as the current frame due to temporal continuity. Hence, utilizing these object queries for mask prediction can provide an effective prior for the model.
We further introduce a Query-Content Interaction Module (QCIM), to make the queries understand the content of current frame, as shown in the bottom of  Fig.~\ref{fig:overview}(c).

In general, our QCIM involves three parts: self-attention (SA), cross-attention (CA) and feed forward network (FFN).
The memory readout will be used in the CA because it has the historical information of all objects.
Let ${X_{sim} \in \mathbb{R}^ {N \times C_v } }$ and ${M \in \mathbb{R}^ {NHW \times C_v}}$  denote the object queries initialized from SIM and serialized memory readout, respectively. The QCIM can be formulated as follows:
\begin{equation}
\begin{aligned}
X^{\prime}&=\mathrm{LN}(\mathrm{SA}(X_{sim})+X_{sim}), \\
X^{\prime \prime}&= \mathrm{LN}(\mathrm{CA}(X^{\prime},M)+ X^{\prime}), \\
X_{qcim}&=\mathrm{LN}(\mathrm{FFN}(X^{\prime \prime})+X^{\prime \prime}),
\label{eq:QCIM}
\end{aligned}
\end{equation}
where $\mathrm{SA}$ and $\mathrm{FFN}$ are utilized to enhance interaction and feature representation of module; $\mathrm{CA}$ is employed to execute query-content interaction and will be elaborated on below.

The main component of QCIM is cross-attention, designed to perform cross interaction between queries and memory readout feature. Formally, it can be described as:
\begin{equation}
\mathrm{CA}(X^{\prime},M)=\mathrm{Softmax}(Q_XK_M^\top)V_M, \\
\label{eq:cross-att}
\end{equation}
where $Q_X$ is the query matrix of $X^{\prime}$; $K_M$ and $V_M$ are the key and value matrices of $M$, respectively. 
Note that during cross-attention, each object query can attend to the content of both its own and other objects, enabling simultaneous content understanding and multi-object interaction.

With the delicately designed SIM and QCIM modules, QMVOS provides scale-aware, object-aware, and content-aware object queries for the model. We utilize these queries as dynamic filters, effectively removing background information and generating precise object masks by dot-producting them with the feature output from the decoder.

\subsection{Implementation Details}
\textbf{Network Details.}
We use ResNet-50~\cite{he2016deep} as the image encoder and ResNet-18~\cite{he2016deep} as the mask encoder. If the current frame should be "memorized", the query feature map will be saved in the memory bank as the memory key, and the image and mask will be channel-wise concatenated and fed into the mask encoder to obtain the memory value. For the decoder, we use the same iterative upsampling architecture to form a feature pyramid network (FPN) as in XMem~\cite{cheng2022xmem}.

\begin{table}[!t]
\setlength{\abovecaptionskip}{0.1cm} 
\centering
\caption{Quantitative comparisons on the Youtube-VOS~\cite{xu2018youtube}.}
\small
{
\begin{tabular}{lccccccc}
\toprule
\rowcolor{mygray}
Method & OF  & M       & \Gsc{}  &  \Jseen{} & \Fseen{} & \Jseen{} & \Funseen{} \\
\midrule
\multicolumn{8}{c}{\emph{YouTube 2018 Validation Split}} \\
\midrule
LWL~\cite{bhat2020learning} & \nomark  & \nomark &  81.5 & 80.4 & 84.9 & 76.4 & 84.4 \\
CFBI~\cite{yang2020collaborative} & \nomark & \nomark &  81.4 & 81.1 & 85.8 & 75.3 & 83.4 \\
SST~\cite{duke2021sstvos} & \nomark & \nomark & 81.7 & 81.2 & - & 76.0 & - \\
CFBI+~\cite{yang2021collaborative} & \nomark & \nomark &  82.0 & 81.2 & 86.0 & 76.2 & 84.6 \\
AOT-L~\cite{yang2021associating}  & \nomark & \yesmark &  81.1 & 81.6 & 86.2 & 74.5 & 82.2 \\
TBD~\cite{cho2022tackling} & \nomark & \nomark & 77.8 & 77.4 & 81.2 & 72.7 & 79.9\\
XMem$^{*}$~\cite{cheng2022xmem} & \nomark & \yesmark & 83.1 & 83.0 & 87.7 & 76.5 & 85.3 \\
\rowcolor{rouse}
\textbf{QMVOS} & \yesmark  &  \yesmark   & \textbf{83.5} & \textbf{83.3} & \textbf{88.0}  & \textbf{77.0} & \textbf{85.8} \\
\midrule
\midrule
\multicolumn{8}{c}{\emph{YouTube 2019 Validation Split}} \\
\midrule
LWL~\cite{bhat2020learning} & \nomark  & \nomark &  81.0 & 79.6 & 83.8 & 76.4 & 84.2 \\
CFBI~\cite{yang2020collaborative}  & \nomark & \nomark &  81.0 & 80.6 & 85.1 & 75.2 & 83.0 \\
SST~\cite{duke2021sstvos} & \nomark & \nomark & 81.8 & 80.9 & - & 76.6 & - \\
AOT-L~\cite{yang2021associating}  & \nomark & \yesmark &  80.8 & 81.0 & 85.5 & 74.4 & 82.2 \\
XMem$^{*}$~\cite{cheng2022xmem} & \nomark & \yesmark & 83.0  & 82.8 & 87.1 & 76.8 & 85.4 \\
\rowcolor{rouse}
\textbf{QMVOS} & \yesmark  &  \yesmark   & \textbf{83.3} & \textbf{83.0} & \textbf{87.4}  & \textbf{77.1} & \textbf{85.8} \\
\bottomrule
\end{tabular}
\label{tab:sota-youtube}
}
\end{table}

\noindent \textbf{Training and Inference.} Our model is trained on video datasets DAVIS 2017~\cite{pont20172017} and YouTubeVOS~\cite{xu2018youtube}. We randomly select eight continuous frames from a video to train the model, following~\cite{cheng2022xmem}. We use the AdamW~\cite{loshchilov2017decoupled} as optimizer with a learning rate of 1e-5 and train the model for 65K iterations with a batch size of 16. For inference, we sequentially segment each frame and store the image and segmentation results every five frames into the memory network.

\begin{figure*}[!t]
\centering
\includegraphics[scale=0.65]{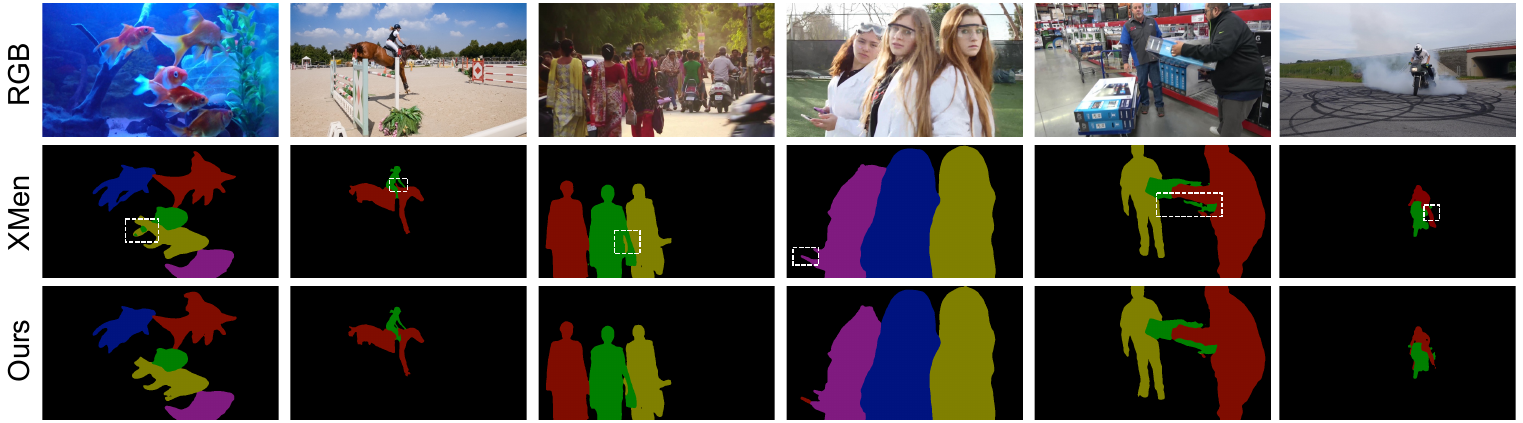}
\caption{Qualitative comparisons of our QMVOS with the SOTA memory-based work, XMem~\cite{cheng2022xmem}. We mark  failures in the white dashed boxes. Our model outperforms XMem in terms of detailing and discriminating similarities.}
\label{fig:img_cmp}
\end{figure*}

\begin{table*}[!t]
	\centering
	\caption{Ablation study. The experiments are conducted on the validation 2017 split of DAVIS~\cite{pont20172017}.}
 \label{tab:ablation}

\begin{subtable}{.3\textwidth}
\center
\caption{Inference speed comparison between baseline method and our method.}\label{tab:speed}
\setlength{\tabcolsep}{2pt}
\small
\begin{tabular}{l c c c c}
\toprule
\rowcolor{mygray}
Method & FPS    & \JandF{}  &  \J{} & \F{}  \\
\midrule
XMem & \textbf{38.1} &  83.4 & 80.5 & 88.3 \\
\rowcolor{rouse}
\textbf{Ours}  &   37.6  & \textbf{84.9} & \textbf{81.4} & \textbf{88.3} \\
    \bottomrule
\end{tabular}
\end{subtable}
\begin{subtable}{.3\textwidth}
\center
\caption{Ablation on initialization.}\label{tab:initialization}
\setlength{\tabcolsep}{2pt}
\small
\begin{tabular}{l c c c}
\toprule
\rowcolor{mygray}
Method &     \JandF{}  &  \J{} & \F{}  \\
\midrule
w/ [$F_{16}$ or $F_{8}$] & \multicolumn{3}{c}{{unable to converge}}  \\
w/ [$F_{16}$, $F_{8}$, $F_{4}$] &   84.6 & 81.2 & 88.0 \\
\rowcolor{rouse}
\textbf{w/ [$F_{16}$, $F_{8}$]}  &   \textbf{84.9} & \textbf{81.4} & \textbf{88.3} \\
    \bottomrule
\end{tabular}
\end{subtable}
\begin{subtable}{.3\textwidth}
\center
\caption{Ablation on the multi-object interaction function in SIM.}\label{tab:interaction}
\setlength{\tabcolsep}{2pt}
\small
\begin{tabular}{l c c c}
\toprule
\rowcolor{mygray}
Strategy &     \JandF{}  &  \J{} & \F{}  \\
\midrule
w/o interaction &   84.5 & 81.3 & 87.6 \\
\rowcolor{rouse}
\textbf{Our module}  &   \textbf{84.9} & \textbf{81.4} & \textbf{88.3} \\
    \bottomrule
\end{tabular}
\end{subtable}

\begin{subtable}{.3\textwidth}
\vspace{0.25cm}
\center
\caption{Ablation on the cross feature.}\label{tab:cross_feature}
\setlength{\tabcolsep}{2pt}
\small
\begin{tabular}{l c c c}
\toprule
\rowcolor{mygray}
Method &     \JandF{}  &  \J{} & \F{}  \\
\midrule
cross with $F_{16}$ feature &   84.3 & 81.0 & 87.6 \\
\rowcolor{rouse}
\textbf{cross with readout}  &   \textbf{84.9} & \textbf{81.4} & \textbf{88.3} \\
    \bottomrule
\end{tabular}
\end{subtable}
\begin{subtable}{.3\textwidth}
\vspace{0.25cm}
\center
\caption{Training strategy.}\label{tab:strategy}
\setlength{\tabcolsep}{2pt}
\small
\begin{tabular}{l c c c}
\toprule
\rowcolor{mygray}
Strategy &     \JandF{}  &  \J{} & \F{}  \\
\midrule
using query from the first frame &   84.5 & 81.2 & 87.8 \\
\rowcolor{rouse}
\textbf{query propagation frame by frame}  &   \textbf{84.9} & \textbf{81.4} & \textbf{88.3} \\
    \bottomrule
\end{tabular}
\end{subtable}
\end{table*}

\section{experiments}
\label{sec:exp}

\subsection{Datasets and Evaluation Metrics}

We evaluate QMVOS on three standard benchmarks: DAVIS 2016~\cite{perazzi2016benchmark} for single-object segmentation, DAVIS 2017~\cite{pont20172017} and YouTube-VOS~\cite{xu2018youtube} for multi-object segmentation. 

For evaluation, we use standard performance metrics: Jaccard index \J{}, contour accuracy \F{} and their average value \JandF{}. 
For YouTubeVOS~\cite{xu2018youtube}, \J{} and \F{} are computed for ``seen'' and ``unseen'' classes separately, and are distinguished by subscripts.
Their average is also reported, denoted as \Gsc{}.
All results are evaluated on official evaluation servers or official tools. 

\subsection{Main Results}

\textbf{Quantitative results.} 
Table~\ref{tab:sota-davis} and Table~\ref{tab:sota-youtube} report the performance of the SOTA methods on the standard VOS benchmarks.
For a fair comparison, we retrain XMem~\cite{cheng2022xmem} using the same configuration as our approach.
We can observe that our QMVOS achieves top-ranked performance on both DAVIS datasets. Notably, our method significantly improves the performance of the baseline method XMem by 1.5 \JandF{} on DAVIS 2017. On the more challenging benchmark Youtube-VOS, our method also achieves optimal performance. These results fully validate the effectiveness of our method.

\noindent \textbf{Qualitative Results.} We visualize some qualitative results in Fig.~\ref{fig:img_cmp} and compare our method with XMem~\cite{cheng2022xmem}. We can observe that our model significantly outperforms XMem in multi-object segmentation scenarios and similar object segmentation scenarios. These comparison results demonstrate the effectiveness of multi-object interaction and object-level perception of our method, respectively.

\subsection{Ablation Study}

To evaluate the effectiveness of the proposed modules, we conduct detailed ablation studies on the DAVIS 2017 val~\cite{pont20172017}.

\noindent \textbf{Impact on inference speed.} As shown in Table~\ref{tab:speed}, we conduct an inference efficiency comparison with the baseline method XMem.
As we operate on sparse queries, the impact on model efficiency is minimal. Our method surpasses XMem significantly while maintaining nearly the same inference speed, thus proving its effectiveness and efficiency.

\noindent \textbf{Effectiveness of multi-scale initialization.} 
Table~\ref{tab:initialization} investigates scale-aware initialization in the Scale-aware Interaction Module (SIM).
With single scale feature for initialization, the model fails to converge. The reason may be that the semantic information provided by single-scale feature is too weak. 
Additionally, incorporating larger scale features $F_4$ cannot improve performance. We believe this is because shallow features contain too much fine-grained information, causing initialized queries to focus less on the deep object semantics.

\noindent \textbf{Effectiveness of multi-object interaction in SIM.} In SIM, we  perform self-attention on object queries to facilitate  multi-object interaction. We also conduct an ablation on this function, and the results are shown in Table~\ref{tab:interaction}. With the assistance of query interaction, our method can improve by 0.4 \JandF, showing the effectiveness of multi-object interaction.

\noindent \textbf{Effect of interaction with memory readout feature.} The memory readout feature has mask information of object. Thus, interaction with readout feature can help queries differentiate between foreground and background. Object queries can also interact with features of other objects, as described in Section~\ref{sec:QIM}. Table~\ref{tab:cross_feature} verifies these functions by applying QCIM on $F_{16}$ from the backbone. Our method can gain 0.6 \JandF \ \ by this interaction mechanism, which is a non-negligible performance improvement.

\noindent \textbf{Effect of query propagation training strategy.}  
Table~\ref{tab:strategy} compares the impact of using queries from the first frame versus the previous frame for training and inference.
Training with the query propagation strategy enables the model to learn frame-to-frame continuity, and also helps it better handle object deformation.
Therefore, query propagation can improve the model performance by 0.4 \JandF.

\section{Conclusion}

This paper proposes QMVOS, a query-based modulation method that incorporates object-level understanding into memory-based method for improved video object segmentation. In QMVOS, we first introduce a Scale-aware Interaction Module (SIM) to summarize objects into high-level descriptors, referred to ``object queries''. Multi-object interactions are also included in SIM to facilitate capturing relationships between objects. 
Additionally, we propose a Query-Content Interaction Module (QCIM) to make the queries understand the context of frame to be segmented. Finally, we filter the features to generate mask using these adaptive queries,  achieving object-level dynamic prediction.
Extensive experiments conducted on SVOS benchmarks demonstrate the effectiveness of the proposed method.

\noindent \textbf{Limitation and Future Direction.} Since query propagation relies on frame-to-frame continuity of the video, our method has limitations in utilizing static images for pre-training. We will explore more flexible query modulation methods to address this limitation in future work.

{
\footnotesize
\noindent\textbf{Acknowledgments:} This work was supported by the National Key R\&D Program of China 505 (Grant No.2020AAA0108303), the National Natural Science
Foundation of China (No. 62192712), the Shenzhen Science and Technology Project (Grant No.JCYJ20200109143041798) and Shenzhen Stable Supporting Program (WDZC20200820200655001). 
}

\scriptsize

\bibliographystyle{IEEEbib}
\bibliography{icme2023template}

\end{document}